## RESEARCH ARTICLE

# Motion Prediction With Gaussian Processes for Safe Human–Robot Interaction in Virtual Environments

**STANLEY MUGISHA**[1,2]**, VAMSI KRISHNA GUDA**[3]**, CHRISTINE CHEVALLEREAU**[4]**, DAMIEN CHABLAT**[4]**, AND MATTEO ZOPPI**[5]**, (Member, IEEE)**

[1]School of Engineering and Technology, Soroti University, Soroti, Uganda
[2]Intelligent Materials and Systems (IMS) Laboratory, Institute of Technology, University of Tartu, 50411 Tartu, Estonia
[3]Institut des Systèmes Intelligents et de Robotique (ISIR), Université Pierre-et-Marie-Curie, 75005 Paris, France
[4]Laboratoire des Sciences du Numérique de Nantes (LS2N), École Centrale de Nantes, 44321 Nantes, France
[5]Dipartimento di ingegneria meccanica, energetica, gestionale e dei trasporti (DIME), Università di Genova, 16145 Genova, Italy

Corresponding author: Matteo Zoppi (matteo.zoppi@unige.it)

This work was supported in part by the University of Genova, and in part by the Centre National de la Recherche Scientifique (CNRS) under the framework of the LobbyBot project: Novel encountered type haptic devices, under Grant ANR-17-CE33.

**ABSTRACT** Humans use collaborative robots as tools for accomplishing various tasks. The interaction between humans and robots happens in tight shared workspaces. However, these machines must be safe to operate alongside humans to minimize the risk of accidental collisions. Ensuring safety imposes many constraints, such as reduced torque and velocity limits during operation, thus increasing the time to accomplish many tasks. However, for applications such as using collaborative robots as haptic interfaces with intermittent contacts for virtual reality applications, speed limitations result in poor user experiences. This research aims to improve the efficiency of a collaborative robot while improving the safety of the human user. We used Gaussian process models to predict human hand motion and developed strategies for human intention detection based on hand motion and gaze to improve the time for the robot and human security in a virtual environment. We then studied the effect of prediction. Results from comparisons show that the prediction models improved the robot time by 3% and safety by 17%. When used alongside gaze, prediction with Gaussian process models resulted in an improvement of the robot time by 2% and the safety by 13%.

## I. INTRODUCTION

Robots have been helpful to humans in many contexts, such as manufacturing, gaming, and health. For tasks where humans work closely with robots maintaining human safety is critical [1]. Collaborative robots (cobots) are designed to maximise human safety, with several measures taken to ensure that the robot does not harm the human. If a collision occurs, the impact on the human should be minimal without unnecessary restarts after a safety stop [2]. In most cases, speed is sacrificed to improve safety. However, this limits the device's efficiency because it is considered too slow. For virtual environments where robots can be used as encountered display devices with intermittent contacts, the efficiency of a device while maintaining user safety is crucial. The speed limitations due to safety constraints behind the design of many cobots have limited their use in immersive virtual environments. While using cobots as encountered type haptic devices in immersive VR to simulate touch with virtual objects, contact between humans and robots is inevitable. This is because the user cannot see the physical robot due to the head-mounted display (HMD) on his head but can only see virtually rendered objects. He/she only comes into contact with the robot at the point of interaction with the objects in the virtual environment (VE) as illustrated in fig. 1 and 2. To simulate interaction with a virtual object as if it were a physical environment, the robot has to be at the point of interaction at the exact time of the physics rendering in the graphics animation software. In addition, the VE is rich and dynamic, with many objects in several positions and oriented differently. Simulating such environments with a robotic device to achieve realistic interaction amidst safety and speed constraints is still a challenge [3], [4], [5], [6], [7]. Due to speed restrictions, the device takes longer to reach some targets, which causes visual-haptic illusions (the differences between what the user perceives and what they can see). As a result, there is a position and orientation mismatch and lag between the virtual object and the haptic proxy. The latency negatively impacts the user experience. Improving device hardware can solve the above challenges. However, safety and complexity considerations frequently result in design choices that make speed limits inevitable. Researchers have proposed different software-based approaches to improve the robot's speed and response. These approaches make use of human intention detection through prediction and velocity modulation [5], [8], [9], [10], [11], [12], [13], [14], [15], [16], [17], [18]. Velocity modulation techniques partition the workspace into a human zone and outer space. The robot's velocity is adjusted according to the location of the robot end effector/tooltip with respect to the human such that when the end effector is considered to be in the outer zone at a greater distance from the human, the robot is moved with a higher velocity. The velocity then is reduced when the robot tooltip is closer to the human in the human zone. Motion prediction predetermines the virtual object location the user wants to interact with, the hand position and the obstacles along the path. With this information, the robot is moved safely to the desired target location.

Researchers have used multi-modal interfaces integrating gaze for human intention detection to improve efficiency and safety further. They have used eye-hand coordination to improve the robot's response based on the fact that eye gaze precedes hand motion for most tasks that involve object manipulation. So, by capturing eye gaze data, different models have been developed to improve human intention detection in human-robot collaboration. For example, establishing how implicit cues such as gaze and head movements precede an action performed by an artificial agent are interpreted may be crucial to improve communication in robot-to-human handover tasks [19], [20]. Ensuring safety without stopping the robot requires motion planning to avoid collisions of the robot against itself and obstacles in the environment, such as humans. Available methods to achieve safe motion plans are slow because generating and executing an optimal collision-free plan takes a long time. Therefore, several measures have to be taken to reduce the robot time and maximize safety. Achieving both simultaneously is still a challenge this research addresses. In this article, we combine velocity modulation, motion prediction, obstacle avoidance, and human intention detection by integrating gaze and hand motion to develop techniques to address efficiency and safety challenges. We designed different motion strategies for a cobot and analyzed their efficiency and safety based on a use case of a virtual reality (VR) application and to analyze the texture of the material to be used in the design an automobile interior [10], [21].

This paper improves the state of the art and contributes to the scientific knowledge through

1) We introduced a data-driven Gaussian process model for online training and prediction of hand motion trajectory from real-time sensor measurements.
2) Evaluation of different methods of kernel computation for online prediction.
3) Human intention detection and prediction strategies using a combination of hand motion and eye gaze.

The rest of the paper is organized as follows: A related works section reviews the state of the art in motion prediction. A methods section which describes the Gaussian process prediction models, related challenges, context of the study and the proposed model. The experiments section presents the evaluation of the models and the prediction strategies. Another section on the experimental setup and evaluation of the strategies, then the results section which presents the results of the analyses and finally a section for the discussion and future perspectives.

## II. RELATED WORK

In human-robot interaction, the problem of the efficiency of robotic devices stems from speed limitations to minimize the impact of accidental collision with objects in the robot's dextrous workspace. However, it is computationally complex to generate collision-free trajectories. Different researchers have tried to circumvent the challenges of speed limitation by attempting to predict human actions and intentions and then planning the robot motions accordingly. By capturing human hand motion and gaze data, prediction models have been developed to generate predictions and robot motions to improve device speeds. Several prediction methods have been proposed. These can be divided into model-based and data-based. Namiki et al. [16] used a minimum jerk model and a particle filter to estimate the operator's intention and predict arm trajectories towards different objects. The minimum jerk model for human motion prediction required recording the initial part of at least half the length of the hand trajectory and then fitting the parameters, including the final position and time. However, human hand motion is non-linear, stochastic, and varies across individuals; therefore, complex to model and estimate. Such models face high inaccuracies, as observed by [22]. To address the above-mentioned challenges, different data-based techniques have been used by various researchers to learn and predict hand motion. The most commonly used approaches depend on neural networks, such as convolution neural networks for data from vision-based systems and recurrent networks if data is from non-vision motion capture systems. The main advantage of the neural network is the ability to capture non-linear dynamics of the hand motion. Landi et al. [8] augmented a minimum jerk with neural networks to get an improved model and then trained on a dataset of human demonstrations to anticipate goal locations in a given workspace [8], [17]. Li et al. [14] used neural networks to model the non-linearity and uncertainty of human hand motion and Bayesian inference to predict motion trajectory by combining early partial trajectory classification and human motion regression. However, the main disadvantage of neural networks is the need for a lot of training data and long training time. Different human intention approaches that rely on a limited amount of training data have been proposed to overcome this scenario. Callens et al. [13] leveraged the power of principal component analysis to recognize and forecast human motion, using a motion detection database of multiple motion models and an estimate of a motion's execution speed. With data from a 3D camera, Ding et al. [9] employed Hidden Markov Models and probability density functions to explain human arm movements and predict user-occupied workspace zones. Bayesian models were used in [15] to infer the hand target from a 3D camera sensor. Other researchers have used models that learn human motions by fusion of human demonstrations with dynamic motion primitives and the Gaussian mixture model [23]. Recently, Gaussian process regression models (GPs) have been used for trajectory prediction. The advantage of GPs as predictors is that they make accurate and smooth predictions with little data without affecting long-term prediction accuracy compared to alternatives like neural networks [24], [25]. In addition, they associate each prediction with an estimate of its uncertainty. GPs have produced good prediction results in previous studies. They have been used to learn from human demonstrations to predict human postures by predicting the human joint velocity given the current posture and robot end-effector velocity [18]. Similarly, in another study [23], a framework that combines partial trajectory classification to recognize human actions with human motion regression using a Gaussian process to predict the trajectory produced good results. However, the main disadvantage of the models is that prediction happens when part of the hand trajectory has been completed. In another study, a multivariate GP was used to construct regression models to reflect the human intentions with respect to the target position [26]. In this model, the GP regression was used to enhance the ability of Dynamic Motion Primitive models in multiple trajectories learning.

The main characteristic of these models is that they were trained offline on a specific task, and the learning is based on emulating a large observation of point-to-point trajectories/motions. In addition, such models cannot be easily transferred to another domain because the performance degrades when the model is presented with new data. Therefore, its use is limited to scenarios where data is similar to the training set. So, they may not be used in dynamic and rapidly changing environments. However, for most real-world problems, (i) the environment is dynamic (changes over time). Therefore, learning must be adaptive and online, (ii) learning samples are few or costly, and (iii) computational resources are limited. A solution would be to have a model that predicts without prior knowledge of the task, easy to train on small samples in a short time, or requires no training at all. Such approaches have been used in [5], [10], and [12]. The models rely on the nearest neighbor metric to determine a target close to the hand. However, the main disadvantage is that only the current hand position or gaze direction is required to predict the final target, and they do not use the trajectory information. To improve the robot's efficiency, predicting a trajectory would enhance the models.

Due to the high refresh rates required for VR applications, online regression with GPs would be suitable for the task of trajectory prediction. However, in the previous studies, the GPs were trained offline with pre recorded trajectories and then inference online due to computational costs associated with training. However, this is not suitable for dynamic environments such as human interaction in VR. This research aims to have a GP model that can be trained online, infer predictions for unseen data at future timesteps in real-time, and then predict the target points.

## III. METHODS
### A. GAUSSIAN PROCESS
A Gaussian process was used to model human motion because it is a non-parametric tool for learning from sample data. The key attribute is the ability to provide uncertainty over estimates [24]. A GP represents distributions over functions and it is specified by the mean and covariance function. GPs are used to infer or predict functions at a finite set of prediction points from observed data.

Given a set of training data $\mathbf{D} = (\mathbf{X}, \mathbf{Y})$ from a noisy sensor, $\mathbf{X}$ and $\mathbf{Y}$ are vectors of input and output data respectively such that $\mathbf{X} = (x_1, x_2, x_3, ..x_n)$ and $\mathbf{Y} = (y_1, y_2, y_3, ..y_n)$.

$$y_i = f(x_i) + w \tag{1}$$

where $w$ is the zero mean, additive gaussian noise with variance $\sigma_n^2$. A GP defines a Gaussian predictive distribution over the output $y_*$ conditioned on training data $\mathbf{D}$ and a test input $x_*$. The GP has a mean

$$GP_\mu(x_*, \mathbf{D}) = \mathbf{K}_*^T (\mathbf{K} + \sigma_m^2 I)^{-1} y \tag{2}$$

and variance

$$GP_\Sigma(x_*, \mathbf{D}) = k_{**} - \mathbf{K}_*^T (\mathbf{K} + \sigma_m^2 I)^{-1} \mathbf{K}_* \tag{3}$$

where, $K$ is the covariance matrix $\mathbf{K}(\mathbf{X}, \mathbf{X})$, formed by the full training data using the kernel function, $\mathbf{K}_*$ is the covariance matrix, $\mathbf{K}(\mathbf{X}, x_*)$ between the test point and the full training data, $k_{**}$ is the kernel function evaluated at the test point, $k(x_*, x_*)$, $m$ is the number of data points, $\sigma_m^2$ is the noise variability, $y$ is the observed output, which also represents the observed underlying function values plus some noise.

The choice of the kernel function depends on the application. For this use case, we used a half-integer Mattern covariance function given as

$$k(x_i, x_j) = \sigma_m^2 \left(1 + \frac{\sqrt{3}d(x_i, x_j)}{l}\right) \exp\left(-\frac{\sqrt{3}d(x_i, x_j)}{l}\right) \tag{4}$$

where $d(\cdot, \cdot)$ is the euclidean distance between 2 points, $l$ and $\sigma$ are the magnitude and length scale hyper-parameters that control the correlation scale and noise variability of the process.

The hyper-parameters $\beta = \{\sigma_m, l\}$ were learned by maximizing the log-likelihood of the training inputs as in eq (5).

$$\beta_{max} = \arg\max_{\beta}(\log(p(y|X, \beta))) \quad (5)$$

### B. CHALLENGES OF GP MODELS

Gaussian processes have been used as the preferred tool for many statistical inference or decision theory problems in machine learning, where flexibility for modeling continuous functions is required. This is because they require few training inputs to produce good results. However, the training takes a long time due to the matrix inversion computation time and the covariance matrix's determinant calculation. This limits their practical application in real-time scenarios which require online training and inference [27].

For a large $m$, the matrix inversion of $(\mathbf{K} + \sigma_m^2 \mathbf{I})$ in eq (2) and (3) is computationally expensive. It requires $O(m^3 n)$ computations where $m$ is the number of data points, and $n$ is the dimensionality of the data. Different approaches have been proposed to overcome the computational overhead of the matrix inversion of the GP. Such approaches include sparse approximation of GPs as used in [28]. The approach constructs a smaller GP representation using pseudo-inputs at different locations. The complexity was reduced to $D \gg m$, where $D$ is the number of real pseudo-data points which is far less than the number of the original data points $m$, and hence obtain a sparse regression method which has $O(D^2 mn)$ training cost. However, the quality of the prediction depends on the number and location of pseudo points, which is a challenge to determine. Another approach is the Infinite Horizon GP (IHGP) [29], which used state space modeling (Kalman filtering) to represent kernels of GPs, suitable for long datasets. The IHGP decreased the computational complexity to linear in the number of data points $O(m^2 n)$. However, this approach is only for equally spaced data points and fails in cases of missing data, such as sensor failures. An improvement to this was the Multi-output Infinite Horizon Gaussian Process (MOIHGP) [30]. It extended the IHGP to cater for unequally spaced inputs and multiple outputs. However, the main problem is that it's a single-input GP and may not exploit the structure of multi-dimensional inputs where the inputs are dependent. In another approach, [27] proposed a different method of matrix decomposition using a high-order lower-rank decomposition (HOLRD) algorithm. They used a direct method for the rapid inversion of this matrix which performs the computation in $O(m \log^2 n)$ by hierarchically factoring the matrix into a product of block low-rank updates of the identity matrix. The HOLRD algorithm is computationally efficient, so we adapted this decomposition as a regression model and extended it to suit our use case. We build on the formulation to construct a multi-dimensional GP suitable for online training and inference. We then used it to design and implement prediction frameworks for human motion prediction to produce robot motions suitable for human interaction in a virtual environment and use the robot as a haptic interface with intermittent contacts.

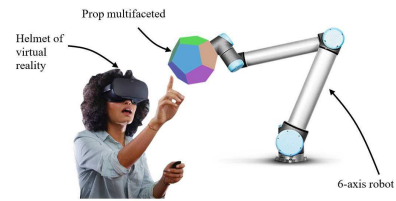

**FIGURE 1.** The experimental platform scheme depicts a user wearing an HMD and touching a prop carried by a cobot.

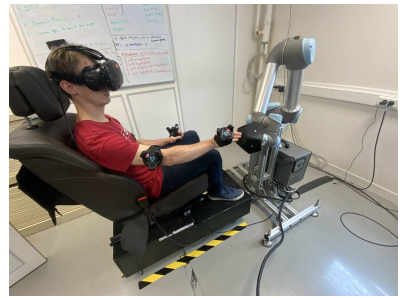

**FIGURE 2.** User interaction in the VR environment.

### C. DESCRIPTION OF CONTEXT

The study assesses the perceived quality of materials in car interiors throughout the first design phases. The application utilized a UR5 robot manipulator. A six-sided prop with multiple texture specimens attaches to the end effector, allowing a user's finger or hand to make contact with the simulated car inside. The user wears an HMD and sits in the actual world to visualise the virtual car (see fig. 2). See [21] and [31] for more details on the configuration of this study application. As explained in [32] and [33], the robot must position and orient the prop with the required side which has a sample of the material associated with the virtual surface, to provide a sense of touching the object.

### D. THE MOTION PREDICTION PROBLEM

The regions of interest for user interaction inside the virtual car are represented by points 1 up to 18, while safe points are numbered from 20-24, as shown in Fig 4. see [10] for details.

Since the human is putting on a VR headset, he cannot see the robot's location. For safety, we implemented trajectory planning techniques to avoid unwanted interactions and collisions between the robot and the human and in addition, maintain a low speed. A virtual sphere in fig 3 was modeled as an obstacle to estimate the user's position and give the system a model to plan safe motions in the Robot Operating System (ROS). So in light of these constraints, the robot must safely arrive at the desired target before the hand to minimize visual-haptic illusions. Prediction, velocity modulation, and safety strategies are employed to solve these challenges.

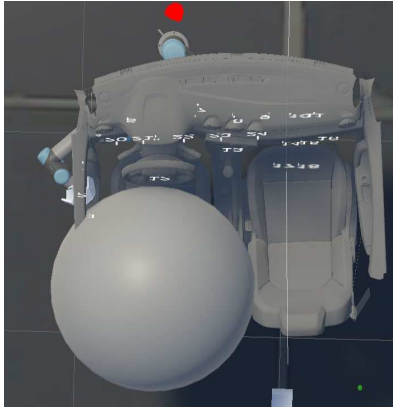

**FIGURE 3.** The Virtual car cockpit.

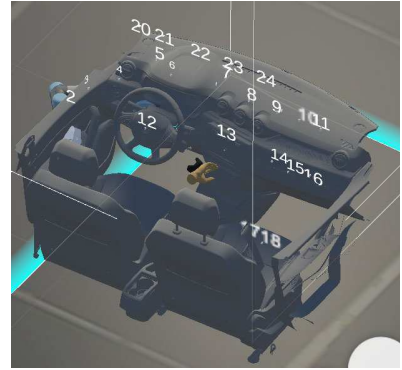

**FIGURE 4.** The interior of a virtual car showing the points 1 to 18 and the safe points 20 to 24.

The human is not restrained and, therefore, can freely move his hand from one point to another as he wishes. Fig 4 shows a snapshot of a hand model moving from one point to another inside a virtual car. The robot must be at the desired point with the orientation of the prop presenting the surface the human user desires to touch in the shortest time securely. This raises the following research problems to address:

- User safety. Motion planning with collision avoidance presented in [34] was used to have a secure system.
- The robot has to be efficient to minimize visio haptic illusions. By design, the robot is slow to keep humans safe and minimize injury in case of a collision. To overcome this, the workspace was partitioned, and velocity modulation techniques were used to move the robot at different velocity profiles depending on its location as described in [10].
- Due to limitations of motion planning, such as time to generate and execute a feasible plan, the robot was slow to respond to changes in human intentions. To improve the response, the points inside the car were fixed; then we generated offline trajectories and recorded them for all possible point-to-point motions.
- It is challenging to predict the whole hand motion trajectory. The hand motion must be tracked, and only a window of the hand motion is used to predict future time steps. A GP model for regression was implemented.
- Since user interaction is free and unscripted, determining which point the user wishes to interact with is a challenge. Different prediction strategies using the nearest neighbor algorithm and the GP were proposed and applied to hand motion and gaze was used to improve efficiency and user safety.

The regions of interest (ROI) in the interior of a virtual car cockpit were determined by a clustering method, selected and designated as points 1-18. To improve safety, safe points were added on top of the dashboard and numbered from 20-24 as shown in fig 4. We used a plane to partition the workspace into two regions; the car interior also referred to as the user workspace and the exterior referred to as the free space (FS) to ensure different velocity profiles for the robot motion. Humans will not enter the FS because it is outside the car in the virtual world, so the robot can move faster. The safe points were placed on a plane to act as the via points for velocity modulation such that the robot moves with a high velocity in the free space and low velocity for the regions inside the car to ensure human safety. The details are explained in [10].

### E. THE PROPOSED PREDICTION MODEL

The motion prediction is modeled as a regression problem. A GP regressor is trained online using a history of hand trajectory data to produce predictions of the hand position at future time steps. A window of time steps determines the length of the input trajectory data, and then a prediction is made at a future time step whose size is determined by a horizon $h$. The model takes in as input a training dataset which consists of the observed hand trajectory from the start $\mathbf{X}_0$ to $\mathbf{X}_t$, and outputs the predicted values $Y$ at any desired time step of the horizon $h$. The output at the horizon timesteps is computed as a multi-step ahead time series forecasting using a naive approach [35]. For simplicity of the notation, the input at time $t$ is denoted as $\mathbf{X}_t$ and the predicted output at a horizon $\hat{\mathbf{X}}_{t+h}$. This is illustrated in fig. 5.

Since the model scales poorly with increased dimensions, three independent GPs are created for each input dimension $x, y, z$ and trained independently to keep the computation complexity linear in dimension.

A significant characteristic of all the GP models is that they assume a zero mean prior [24]. The resultant effect is that the predicted mean tends towards zero for longer prediction time steps of the horizon, and the uncertainty increases with distance from an observed value. This makes the model unsuitable for longer regression, where the objective is to predict several time steps. To overcome this problem, we propose a 2-dimensional (2D) GP which uses both the velocity and the hand's position. We shall refer to our model as the Enhanced GP (EGP) for the rest of the document. The main advantage is that the training of the velocity will have small values close to zero and, therefore, will not suffer from the zero-mean prior problem. Another advantage is that the model exploits the dependency relationship between velocity and position. So we can have predictions at longer horizons with fewer errors and uncertainty. Training the model relies on both the velocity and positions, but the regression is done with the velocity and augmented to the predicted positions. The algorithm is explained in algorithm 1 Online GP Predict. The input is a vector of training points whose size is defined by a window. The objective is to output a datapoint at a time-stamp defined by a horizon $h$. We create a 2D array of length $w$ for each axis for the position and velocity. The queue is filled with trajectory points at each time step to fill up the window (ln 3-4). Then a 2D GP is created and initialized with kernel parameters and the system noise. The GP is then trained, and the hyperparameters are optimized to fit the model on the training data (ln 5). Then a set of predictions for the mean position is generated using the training data from $w_0$ to $w_t$. This is the latent space representation from the training data (ln 6). The second dimension of the GP is for the velocity signal. This is used to generate the predictions up to the desired horizon $h$ (ln 7). On line 8, the predicted position is computed from the velocity and concatenated to the mean predictions for the positions. When a new data point is received, the oldest value is removed from the training list to maintain the window size, and the process is repeated at the next sampling period.

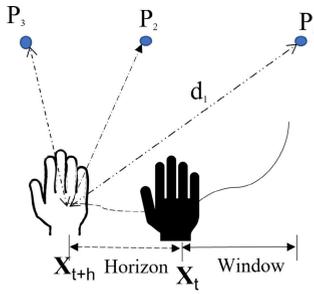

**FIGURE 5.** Motion of hand from point $p_1$ to $p_3$. A solid line shows the hand's trajectory, and the hand's current position at $X_t$ is bold. The predicted trajectory is dotted, the hand's position at a horizon $h$ is $X_{t+h}$, and the hand representation is unshaded. The training window is solid.

## IV. EXPERIMENTS

Four people participated in thsese experiments. The subjects were researchers in the lab, with immersive VR experience ranging from none to moderate use. An HTC Vive Eye® HMD, a tracker attached to the hand trackers, and a car seat were used. The test applications were run on a desktop computer with Unity3D® VR software. The GP models used a mattern kernel with the initial parameters set to 1.0 for the length scale and the noise variance set to 0.003m for position and velocity. The noise variance was selected based on previous studies on HTC Vive tracking accuracy where human motion is involved [36].

---

**Algorithm 1** Online GP Predict

**Require:** Training data $\mathbf{D}$, window size $w$, horizon $h$
**Ensure:** predicted position $\hat{x}_{t+h}$, covariance

1: **function** EGP_Predict($\mathbf{D}, w, h$)
2:     Initialise a 2D GP with hyper-parameters $\beta$
3:     **while** $\mathbf{D} \leq w$ **do**
4:         Add a new data point $\mathbf{X}_t = [x_t, \dot{x}_t]$ to $\mathbf{D}$
5:         Train GP on $\mathbf{D}$, and optimize $\beta$ with eqn. 5
6:         $\hat{x}_t \leftarrow \mu_1 \text{ GP}(\mathbf{D}, w_t)$
7:         $\hat{\dot{x}}_t \leftarrow \mu_2 \text{ GP}(\mathbf{D}, h)$
8:         $\hat{x}_{t+h} \leftarrow \hat{x}_t + \hat{\dot{x}}_t h \delta t$.
9:         remove first data point $\mathbf{D}_o$ from $\mathbf{D}$
10:    **end while**
11:    **return** $\hat{x}_{t+h}$
12: **end function**

---

### A. EVALUATION OF GP ALGORITHMS

One of the key factors for realistic haptic rendering is time. For an immersive experience, the graphics images for VR must be re-calculated at a minimum of 30 frames per second (FPS) [37], [38]. This presents a constraint on the training and prediction time for the algorithm. So, the computational time for training and predictions had to be within these limits. For this experimental study, the VR scene was updated at a rate of 34 FPS, and data was captured from the HTC Vive motion tracking system at a frequency of 34Hz.

We had two primary objectives for the study. The first was to analyze the training time for the different models: the Basic, Holrd and EGP, then select a suitable training window which satisfies the time constraints. The second task was to find a suitable prediction horizon with minimal error and uncertainty.

#### 1) WINDOW SELECTION

We compared the efficiency and model prediction quality for all the implementations. Efficiency was evaluated as the total time for hyper-parameter optimization on training data and prediction on the unseen dataset (test set) and the quality of the model was assessed by the log-likelihood (LL), which is a measure of model fit.

We studied three different GP implementations. The full basic GP proposed by [24] as the baseline, the GP Holrd presented in [27], and our implementation the EGP to produce preliminary results.

The dataset used was sample hand trajectories recorded when the hand moved from point to point in an immersive virtual environment.

We used GP implementations in C++ to achieve better speeds. The hyperparameters were optimized by maximizing the LL in eq (5) using the limited memory Broyden-Fletcher-Goldfarb-Shanno with boundaries (L-BFGS-B) algorithm [39]. Each GP was trained on a window, which consisted of data points from the start of a trajectory,

**TABLE 1.** Time and log-likelihood for algorithm computed at different windows.

| Window (s) | Time (ms) | | | Log Likelihood | |
|---|---|---|---|---|---|
| | Basic | Holrd | EGP | Basic/Holrd | **EGP** |
| 0.5 | 11.66 | 1.09 | 1.58 | 51.14 | **81.70** |
| 1.0 | 20.33 | 1.99 | 2.00 | 131.44 | **161.72** |
| 1.5 | 22.17 | 2.42 | 2.71 | 165.59 | **254.51** |
| 2.0 | 30.04 | 3.82 | 3.35 | 270.20 | **340.23** |
| 2.5 | 36.78 | 3.81 | 4.39 | 326.16 | **433.02** |
| 3.0 | 44.23 | 5.54 | 5.62 | 387.41 | **530.03** |

**TABLE 2.** Mean Error by horizon for the GP algorithms.

| Horizon (%age) | MAPE | | RMSE | |
|---|---|---|---|---|
| | Horld | EGP | Horld | EGP |
| 5 | 3.883 | **2.071** | 0.005 | **0.004** |
| 7.5 | 4.084 | **2.586** | 0.006 | **0.005** |
| 10.0 | 4.371 | **2.920** | 0.007 | **0.007** |
| 12.5 | 4.536 | **3.417** | 0.009 | **0.008** |
| 15.0 | 4.915 | **3.796** | 0.010 | **0.009** |
| 17.5 | 5.390 | **4.253** | 0.012 | **0.010** |
| 20.0 | 5.851 | **4.905** | 0.013 | **0.011** |

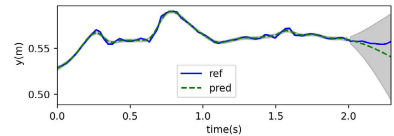

**FIGURE 6.** GP Horld model prediction on the test data set using a window of 2s and a horizon of 15%. The predicted means are shown by a dotted line and the uncertainty is in gray. Notice the uncertainty as the horizon increases. The MAPE in the prediction was 4.46 and the RMSE 0.031.

and evaluated on the next part of the trajectory, which we shall refer to as the horizon.

For this evaluation, we used the recorded trajectories and for each trajectory, a training list $D$ of size $w$ was used as explained in algorithm 1 to compare the GP implementations. This was run 100 times, then the meantime and LL were computed and then tabulated in a row. Then $w$ was varied to obtain more results as summarized in table 1. We used windows of different sample sizes of $w = 17, 34, 51, 68,$ and 85, representing time steps of 0.5s, 1.0s, 1.5s, 2s, and 2.5s respectively. We had a total of 108 samples and results for analysis.

The results showed that a longer window produced a higher value of the LL in all the models. This indicated that the model quality improved with more training data. Overall, the EGP algorithm performed better in terms of hyperparameter optimization. This was evidenced by the highest LL values for each window, while the basic GP and the Holrd others had a similar value. The high value of the LL can be attributed to the fact that the optimization was done on small values of velocity. The other two algorithms had a similar value because they are one dimensional, and the optimization was done on larger values.

For efficiency, results show that time increased with the window size. A longer window was associated with an increase in time due to an increase in the volume of data, which required a longer time for parameter optimization. Overall, the results showed that the Holrd was the most efficient for all windows with the lowest time than the EGP and basic GP. The time for Holrd was slightly shorter than the EGP because it's one-dimensional. The extra velocity dimension in the EGP added more data to the model, which required more time to optimize. The basic GP was one-dimensional but took the most time due to the direct matrix inversion computation.

Given the time constraints of the VR refresh rate, the required time was less than 11ms; comparing the computation time for data in each window showed that the basic GP would not satisfy the time constraint. Results in table 1 show that either the Holrd or EGP could be used.

### 2) HORIZON AND PREDICTION QUALITY

When a trained model from the trajectory is to be used for prediction, the possible learning error and prediction accuracy must be considered to ensure the system's safety.

This objective was to select the best prediction horizon with a small error in the predictions. The evaluation criterion was the Mean Absolute Percentage Error (MAPE) and the Root Mean Squared Error (RMSE) between the predictions and the actual values in the test set. The horizon was selected as a percentage of the training data (window) in a sequence. Such that the data of the horizon was the sequence of data points following the last point of the window as shown in Fig. 5. For the experiment, different prediction horizons were evaluated, corresponding 5%, 7.5%, 10%,12.5%, 15%, 17.5%, and 20% of the window. The MAPE was calculated between the set of all predictions and their corresponding actual values. Only two Holrd and EGP were evaluated.

Overall, the results for the algorithms show that the predictive error increased with an increase in the horizon (see table 2). The reason is the longer prediction horizons associated with increasing distance from the training set; the error grows rapidly. A comparison showed that EGP produced smaller errors than Horld for all the Horizon. This can be attributed to the quality of parameter optimization thanks to the velocity component of the GP in the training data.

From the results, a horizon of 15% with the EGP was determined as the best algorithm because it was the most extended prediction horizon we could achieve within the acceptable error bounds, beyond which the error rate grows faster. Fig. 6 and 7 show the prediction results from the two algorithms.

### B. RESULTS 1: EVALUATION OF GPS
### C. GP PREDICTION FRAMEWORK FOR SAFE HUMAN-ROBOT INTERACTION

After model evaluation, optimal parameters were determined as shown in table 3. These were later used for predicting human motion for online intention detection. The EGP model was used in different strategies and its contribution to user safety and efficiency of the robot was evaluated. This subsection presents the strategies proposed to move the robot between different points.

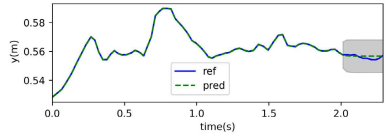

**FIGURE 7.** GP EGP model prediction on test data set using a window of 2s and a horizon of 15%. The predicted means are shown by a dotted line and the uncertainty is in gray. The MAPE in the prediction was 3.24 and the RMSE 0.018.

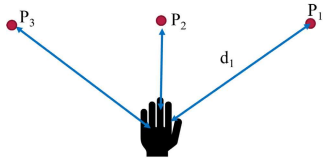

**FIGURE 8.** NN Prediction.

### 1) STRATEGY A: PREDICTION WITH NEAREST NEIGHBOR (NN)

The state of the art was first presented in [5] and used in [10] and [12]. The objective is to select a point nearest to the hand from a list containing all points inside the car (pts 1-18) as shown in Fig 4. At every time stamp, the distance of each point from the hand is calculated, and the point with the shortest distance is returned as the desired point. The strategy is illustrated in fig. 8.

### 2) STRATEGY B: GP-NN PREDICTION

This strategy relies on determining the next point based on future hand position. A GP is used to predict the hand position at a horizon $h$ using algorithm 1. Then the NN algorithm determines the target using the predicted position as illustrated in Fig. 5. This strategy algorithm is described in algorithm 2.

---

**Algorithm 2** STB: GP-NN

**Require:** training data list $\mathbf{D}$, window $w$, horizon $h$, hand position and vel $\mathbf{X}_t$, interior points $\mathbf{P}$

**Ensure:** predicted point $P^*$

1: **function** STB($\mathbf{X}_t, D, w, h, \mathbf{P}$)
2: $\quad \hat{P}_h \leftarrow \mu GP(X_t, D, w, h)$ $\qquad \triangleright$ *Hand pose from GP*
3: $\quad P^* \leftarrow \min\limits_{P_i \in P} \left\| \hat{P}_h - P_i \right\|$ $\qquad \triangleright$ *Point Predicted by GP*
4: $\quad$ **return** $P^*$
5: **end function**

---

### 3) STRATEGY C: SAFE NN

To improve the user's safety and robot time, we use the following:

1) Intention detection. We detect if a human wants to move his hand from one point to another by use of a

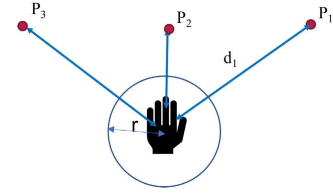

**FIGURE 9.** Safe NN Prediction with threshold.

threshold distance on the hand $r$ such that if any point is within $r$, we know that the user wants to interact with a point, and the robot's motion is restricted. And if the point is outside the sphere, the human wants to move to another point. So another point of interaction is selected.

2) For safety, the robot should move as far as possible from the human. Via points were added outside the human space, and the prediction algorithm was designed to move the robot to the predicted point through the via points. These are referred to as safe points numbered ($SP_{20} \ldots SP_{24}$) and located in the exterior of the car workspace (Figs 3 and 4). Motion through safe points is with a high velocity using velocity modulation techniques presented in [10] and [11].

This strategy is summarized in algorithm 3, the desired point is the nearest to the hand within a sphere of radius $r$. A safe point associated with the nearest point is taken if no such point exists. The algorithm inputs the hand position, the points of interaction, and the safe points, then outputs a desired point. An example in fig. 9 illustrates the strategy. The point $P_2$ is the closest to the hand and therefore chosen. If a point is at a distance of more than the threshold from the hand, the robot will go to a safe point $SP_i \in SP$, which is the closest to $P_2$.

---

**Algorithm 3** Strategy STC: Safe NN

**Require:** Hand position $P_h \in \mathbb{R}^3$, Hand threshold $r$, points $\mathbf{P}$. (shown in Fig. 9).

**Ensure:** predicted point $P^*$.

1: **function** STC($P_h, \mathbf{P}, r$)
2: $\quad P \leftarrow \min\limits_{P_i \in \mathbf{P}} \|P_h - P_i\|$ $\qquad \triangleright$ *Point Predicted by NN*
3: $\quad$ **if** $d(P, P_h) < r$ **then**
4: $\qquad P^* \leftarrow P$
5: $\quad$ **else**
6: $\qquad P^* \leftarrow \min\limits_{SP_i \in SP} \|P_h - SP_i\|$ $\qquad \triangleright$ *SP selection*
7: $\quad$ **end if**
8: $\quad$ **return** $P^*$
9: **end function**

---

### 4) STRATEGY D: SAFE GP-NN

The GP prediction model offers two benefits. First, with the knowledge of the future hand position, we can move the robot to a safe point and take advantage of velocity modulation. Secondly, we can use the predicted hand motion to move the robot to a close point earlier than in Strategy C.

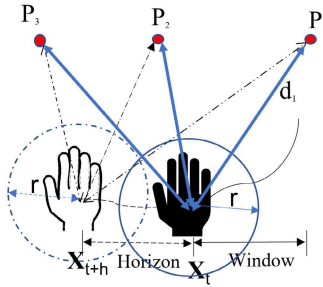

**FIGURE 10.** Safe GP NN Prediction with threshold.

However, we noted that the predicted hand position has much uncertainty. The predictions should be used only when the real hand position is far from any of the points. This is because there is negligible advantage for GP prediction if the real hand is close to a point. The strategy switches between the predicted and the real hand positions using a threshold distance $r$ such that the result is either one of the points selected by the real hand position using NN or the predicted position by a GP as illustrated in fig. 10. The result from the predicted hand position is either a safe point or an interior point. To improve safety, a scaling factor $\alpha$ was introduced to weigh the safe points detection more than the interior points, thus giving preference for the robot to move through safe points rather than points inside the car. $r_1$ is the scaled distance of the real hand from the closest point while $r_2$ is the scaled distance of the predicted hand position from the closest point as shown in lines 5 and 6 in algorithm 4. Switching to select a point by the real hand position or the predicted position is done using $r$ such that if $r_1 < r$, we take a point selected using the real hand position (ln 7-8). Otherwise, we check if $r_2 < r$, then take an interior point selected by the predicted hand position or a safe point (ln 10-14).

### 5) STRATEGY E: SAFE NN WITH GAZE
Hand-to-eye coordination was used to determine human intention and improve selection efficiency. This strategy and the related algorithm was introduced in [10]. Here, the eye gaze was used to preselect points of interaction. Since the hand is the primary contact with the points, it is then used for finer selection such that if a point is within the radius $r$, it is selected as the desired. Otherwise, a safe point in the gaze direction is selected. An illustration is shown in Fig. 11.

### 6) STRATEGY F: SAFE GP-NN WITH GAZE PREDICTION
To improve the selection of strategy E, we try to predict the hand motion after preselection by the gaze. We follow the same procedure as in strategy D. The only difference in this algorithm is that we have another added selection. So the selection is first by gaze direction, then the actual hand, then the predicted hand position. This is illustrated in Fig. 12.

---

**Algorithm 4** Strategy STD(EGP): Safe GP NN

**Require:** hand state $\mathbf{X}_t \in \mathbb{R}^3$, Hand Threshold $r$, window $w$, horizon $h$, points $\mathbf{P}$, $\alpha$. (Shown in Fig. 12).
**Ensure:** predicted point $P^*$.

1: **function** STD($\mathbf{X}_t, r, w, h, \mathbf{P}$)
2: $\quad \hat{P}_h \leftarrow \mu GP(\mathbf{X}_t, w, h)$  ▷ *Hand pose from GP*
3: $\quad \hat{P} \leftarrow \min\limits_{P_i \in P} \left\| \hat{P}_h - P_i \right\|$  ▷ *Point Predicted by GP*
4: $\quad P \leftarrow \min\limits_{P_i \in P} \left\| P_h - P_i \right\|$  ▷ *Point Predicted by NN*
5: $\quad r_1 \leftarrow d(P_h, P) * \alpha$
6: $\quad r_2 \leftarrow d(\hat{P}_h, P) * \alpha$
7: $\quad$ **if** $(r_1 < r)$ **then**
8: $\quad\quad P^* \leftarrow P$
9: $\quad$ **else**
10: $\quad\quad$ **if** $r_2 < r$ **then**
11: $\quad\quad\quad P^* \leftarrow \hat{P}$  ▷ *Point Predicted by GP*
12: $\quad\quad$ **else**
13: $\quad\quad\quad P^* \leftarrow \min\limits_{SP_i \in SP} \left\| SP_i - \hat{P} \right\|$  ▷ *SP selection*
14: $\quad\quad$ **end if**
15: $\quad$ **end if**
16: $\quad$ **return** $P^*$
17: **end function**

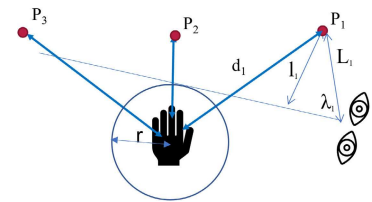

**FIGURE 11.** Safe NN Prediction with a threshold. A point $P_i$ with the minimum $\lambda_i$ is first selected. But if $P_i$ is not within $r$, a safe point is selected.

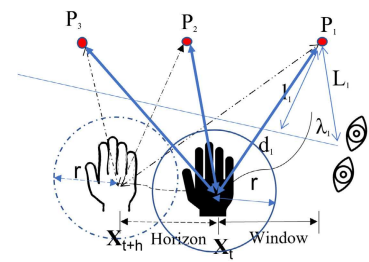

**FIGURE 12.** Safe NN Prediction with Gaze direction.

The algorithm inputs the hand position, the gaze direction, the points of interaction, and the safe points, then outputs a desired point.

A few rules guide our selection as shown in algorithm 5. First is that eye gaze gives a rapid selection of the interest point, then based on a threshold $r$ we decide whether it is necessary to predict the selected point. If the scaled distance between the point selected by gaze $P_{gz}$ and the hand is within a threshold, then $P_{gz}$ is the desired, and the algorithm stops. Otherwise, we have to select a point based on the predicted hand position (ln 9) such that we find the closest point $P_{gp}$ to the predicted hand position $\hat{P}_h$. For the selected $P_{gp}$ if $r_2$ is less than $r$, then $P_{gp}$ is selected as the desired (ln 11), else a safe point is selected.

---

**Algorithm 5** Strategy STF: Gaze Safe GP NN (STF -EGP)

**Require:** Hand State $\mathbf{X}_t \in \mathbb{R}^3$, Hand threshold $r$, window $w$, horizon $h$, Points space $\mathbf{P}$, $\alpha$. (Shown in Fig. 12).
**Ensure:** predicted point $P^*$.

1: **function** STF($X_t, r, w, h, P$)
2:      $\hat{P}_h \leftarrow \mu GP(X_t, w, h)$      ▷ *Hand pose from GP*
3:      $P_{gz} \leftarrow \min\limits_{P_i \in \mathbf{P}} \lambda_i$      ▷ *Point Predicted by gaze*
4:      $P_{gp} \leftarrow \min\limits_{P_i \in P} \left\| \hat{P}_h - P_i \right\|$      ▷ *Point Predicted by GP*
5:      $r_1 \leftarrow d(\hat{P}_h, P_{gz}) * \alpha$
6:      $r_2 \leftarrow d(\hat{P}_h, P_{gp}) * \alpha$
7:      **if** $r_1 < r$ **then**
8:          $P^* \leftarrow P_{gz}$
9:      **else**
10:          **if** $r_2 < r$ **then**
11:              $P^* \leftarrow P_{gp}$
12:          **else**
13:              $P^* \leftarrow \min\limits_{SP_i \in \mathbf{SP}} \left\| SP_i - P_{gp} \right\|$      ▷ *SP selection*
14:          **end if**
15:      **end if**
16:      **return** $P^*$
17: **end function**

---

### D. ROBOT MOTION PLANNING

The motion planning is based on the ROS framework. When a strategy selects a point, the robot computes and executes the trajectory from its current position to the selected target. However, this may entail stopping the robot's motion and recalculating an obstacle-free trajectory online. To save time for calculations, we used predefined offline trajectories with short execution times.

## V. EXPERIMENTAL SET UP AND EVALUATION OF THE STRATEGIES

During the experimental procedure, the optimal window size selected was 2.0s, similar to other GP-related experiments [29], [30]. A horizon of 15% of the window size was selected. Other parameters related to the experiment are listed in table 3.

### A. EXPERIMENTAL SETUP

The participant was seated in the car seat 0.6m above the ground, -0.1m and 0.9m in $x$ and $y$ respectively from the robot base frame while the robot was fixed on a 0.8 m high table. The placement of the robot in the scene was chosen to reach all the places where the user's hand would interact with the robot's probe. A detailed explanation of the user and robot placement is given in [40]. The user wore an HMD, and Vive trackers were attached to the user's hands as shown in Fig. 2.

**TABLE 3.** Parameters used in the experiment for the HRI task.

| Parameter | Value |
|-----------|-------|
| r | 0.2m |
| Vm | 0.4m/s |
| Vr | 0.25m/s |
| $l$ | 0.5 |
| $\sigma_m$ | 0.003 |
| w | 2.0s |
| h | 15% |
| $\alpha$ | 0.8 |

In this experiment, we used seven trajectories. Two of these were long-distance trajectories (from points 2 to 11 and 5 to 18), three were of a medium distance (from points 5 to 11, 5 to 15, and 12 to 15), and two were of short distances (from points 3 to 4 and 17 to 16). Then, for each trajectory, the user was instructed to move his hand from a start point to a defined endpoint. The data from the HMD and the user's dominant hand were recorded to be used for the GP training. The data comprised the position and velocity of the hand, the head position, and orientation. The user was instructed to do the task at the speed he wanted. Throughout the data collection process, the robot was stationary to ensure the security of the user. This was mainly because we wanted to compare the strategies with the same input data. The goal was to predict the trajectory at a horizon and then identify the point of interaction. This problem is regression and then determining the nearest point in a closed set to a specified point (hand position). We used the head-mounted display to identify points in the user's vision and if their gaze is focused towards a specific point $P_i$. We then expressed the distance of the points to the gaze direction as a function of $l_1/L_1$ as shown in fig. 11 and 12. The user would then direct his hand to a location within the virtual car. The goal was to detect as quickly as possible the point the human would likely reach to interact with and then to move the robot to the point.

1) Once the direction is decided, the robot can be moved to intermediate places to simplify the task.
2) Safe-points $SP_i \ \forall i \in [20, 24]$ are defined on the boundary plane outside the car's interior space. Between these points, the robot can move faster.
3) At each time stamp, we define the target point $P_i \ \forall i \in [1, 18]$ for prediction.

### B. EVALUATION CRITERIA

The main objective was to compare our GP implementation against the state of the art regarding improving efficiency and user safety. The comparison is done by carrying out three assessments.
1) Effect of prediction on efficiency.
2) Effect of prediction on safety.
The efficiency evaluation criteria used the following:
1) $T_d$: Time taken by the strategy to detect the desired point the user wants to reach. We calculated it as the

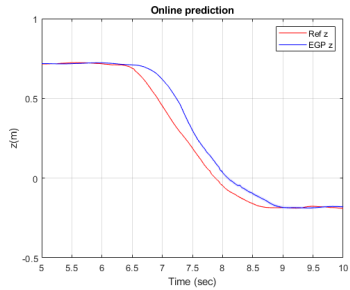

**FIGURE 13.** Prediction results for the EGP compared with the reference signal signal.

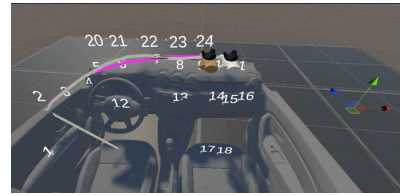

**FIGURE 14.** Hand motion from points 2-11 in unity software for STA and STB. The predicted hand is shown in white with a white trail and the real hand is brown with a pink trail.

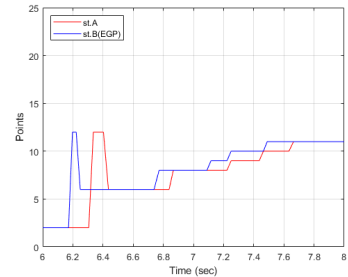

**FIGURE 15.** Comparison results for detection from hand motion by the strategies STA and SSTB(EGP).

**TABLE 4.** Results for efficiency for the three strategies for trajectory 2-11 without safety.

| ST | $T_d$ | $T_r$ | $D_r$ |
|---|---|---|---|
| STA | 7.66 | 12.99 | 1.24 |
| STB(EGP) | **7.46** | **12.85** | 1.24 |

duration from when the trajectory starts to when the desired endpoint is detected.

2) $T_r$: Time taken by the robot to reach the desired/final of the user.

3) $D_r$: Distance traveled by the robot from start to end for all points the robot travels through. The lower the distance, the more efficient the robot is. This criterion can be regarded as a measure of resources in terms of energy consumed to move the robot from point to point.

The safety evaluation criteria used the following:

1) $SP_d$: Number of safe points detected by the strategy. The strategy detects the safe points as the hand moves from the start to the final point.

2) $SP_r$: Number of safe points traveled by the robot. It is the number of safe points from the start to the endpoint the robot uses as via points.

3) $D_h$: Mean distance of human from robot. We calculated from the robot prop to the human sphere. This distance characterizes the safety of the user. The further the robot is from this sphere, the safer the user.

## VI. RESULTS

Results from the comparisons are presented first with an analysis of one trajectory and then a comparison of the seven trajectories.

### A. ANALYSIS OF RESULTS FROM TRAJECTORY 2-11: EFFICIENCY AND SAFETY

#### 1) ONLINE PREDICTION

A comparison of prediction for the EGP explained in algorithm 1 against the reference signal (without prediction was done). The predicted signal was filtered by a moving average of 11 points. A graph for the results of the z-axis is shown in Fig. 13. Results show that the predicted signal is always ahead of the real signal but with high variance since the prediction result is based on unseen data points, which is highly uncertain. The GP uses the learned trajectory to predict a likely point 15% time-steps ahead of the trajectory.

#### 2) DETECTION FOR STRATEGY A AND B

From the hand motion, we compared results for the intention detection using the signals with prediction using STB(EGP), and STA without prediction. The comparison is shown in fig. 15 and table 4. From the results, time for detection was best with STB(EGP) at 7.46s and STA at 7.66s. The number of intermediate points detected was the same. Regarding the time for the robot, the arrival time to the final point was 12.85s and 12.99s for STB(EGP), and STA, respectively. A comparison of the robot motion for the strategies is shown in fig. 17. The hand motion for the predicted and real hand position is shown in fig. 14 and a related video for the corresponding animation.

*Robot Motion:* For STA, the resultant motion is in fig. 16. We can see that the motion from points 2 to 11 has five intermediate points. The detection of points is in the order of 2, 12, 6, 8, 9, 10, 11. The strategy started with point 2, then at time t = 6.3s, point 12 was detected, then point 6 at 6.4s, then point 8 at 6.84s, point 9 at 7.22s, then 10 at 7.43s and finally point 11 at 7.63s. The robot started at point 2, travelled to the first detected point 12 at 6.31s, and arrived at 9.51s. Because the trajectory from 2-11 is long, about 3.2s, it executed this trajectory until completion. So it did not execute the trajectories for the intermediate points detected before completing the first trajectory. After executing the trajectory from 2-12, the robot received information about the trajectory from 12-11, executed it at 9.51s, and finally arrived to point 11 at 13.00s.

Motion with STB(EGP) was similar to STA. Detection and robot motion was through the same points but at different times. Detection started at point 2, then point 12 at 6.17s. The robot, at that moment, started moving to point 12. As the hand progressed, point 6 was detected at 6.22s then 8 at 6.67s, 9 at 7.09s, 10 at 7,22s, and finally point 11 at 12.8s. During this period, the robot was still executing the trajectory from 2-12, so all the trajectories to the intermediate points were discarded. It only executed the new trajectory 12-11 when it had reached point 12 at 9.37s, and then it reached point 11 at 12.85s.

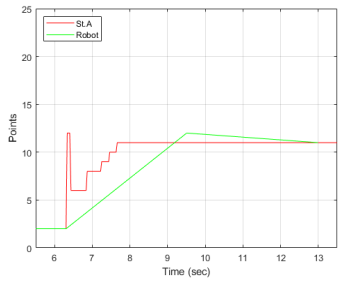

**FIGURE 16.** Robot motion according to strategy STA.

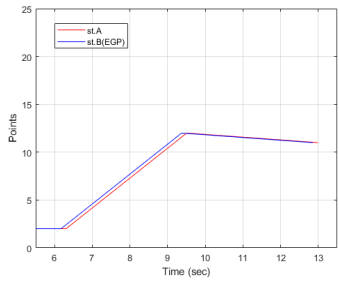

**FIGURE 17.** Robot motion compared for STA and STB.

**TABLE 5.** Results for detection with safety for the trajectory 2-11.

| ST | $T_d$ | $SP_d$ | $T_r$ | $D_r$ |
|---|---|---|---|---|
| STC | **7.87** | 2.00 | 11.37 | 1.38 |
| STD(EGP) | 8.10 | 2.00 | **11.26** | 1.38 |

### 3) DETECTION WITH SAFETY
A comparison for STC and STD(EGP) prediction in table 5 shows that detection with safety was best with STC. STD(EGP) seemed poor at detecting the final point. However, the graph in fig 18 shows detection took more time through the safe points than interior points. The reason is that the STD(EGP) preferred to maximize safety by traveling longer through safe points due to the scaling factor $\alpha$ introduced in algorithm 4. STD(EGP) had the longest time through the safe points and maybe the safest.

### 4) ROBOT MOTION WITH SAFETY COMPARED
The results for detection based on hand motion are shown in fig 18, and the corresponding robot motion is shown in fig 19. We noted that the detection and robot motion travels through safe points before arriving at the final point. The safety results are shown in table 6. The Robot motion was shorter for STD(EGP) because of fast detection while leaving the start point to the next points. The number of safe points and the distance traveled by the robot were the same (see table 5).

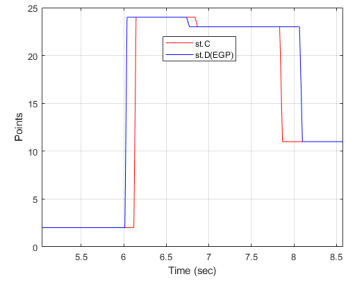

**FIGURE 18.** Comparison results for detection from hand motion by the strategies STC and STD(EGP).

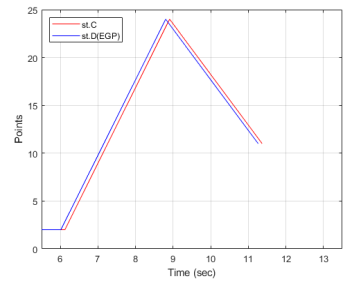

**FIGURE 19.** Comparison results robot motion by the strategies STC and STD(EGP).

**TABLE 6.** Results for safety for the motion of robot for the trajectory 2-11.

| ST | $SP_d$ | $SP_r$ | $D_b$ |
|---|---|---|---|
| STC | 2.00 | 1.00 | 0.95 |
| STD(EGP) | 2.00 | 1.00 | 0.95 |

With STC, Detection started at point 2, then $SP$24 is detected at 6.12s, then $SP$23 at 6.84s, and finally point 11 at 7.83s. For this strategy, the robot started its motion from point 2, traveled to the first detected safepoint $SP$24, executed the trajectory from 2-24, and arrived at $SP$24 at 8.91s. However, the intermediate points were ignored. While at $SP$24, it received the command to move to point 11, and trajectory 24-11 was executed. The robot finally arrived at point 11 at 11.37s. With, STD(EGP), Motion started at point 2, then $SP$24 was detected at 6.01s, then the strategy detected $SP$23 at 6.74, and finally point 11 detected at 8.07s.

*Safety Comparison:* For safety, In all the strategies, the number of safe points detected was 2, while the robot traveled through only one.

### 5) COMPARISON WITH GAZE
*Detection Efficiency:* By including gaze, the number of safe points were detected increased to three as shown in fig. 20. Overall, STE had the best detection time, and then STF(EGP). The late detection in prediction algorithms was mainly enhanced to give preference for safe points such that the robot takes a longer time moving through the exterior of the car. However, the overall distance covered by the robot was the same.

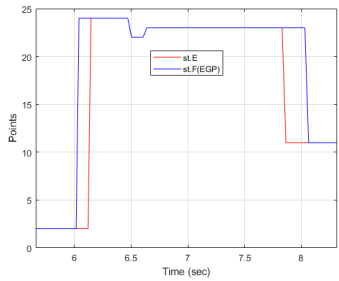

**FIGURE 20.** Detection for strategy STE and STF (EGP).

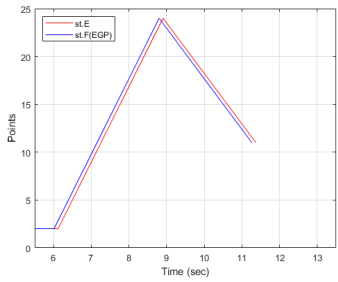

**FIGURE 21.** Robot Motion for strategy STE and STF (EGP).

*Safety:* Results for user safety with gaze for STE, and STF(EGP) were similar.

Generally, the main observation from this analysis showed that prediction improved the robot time and safety. Gaze improved the detection of safe points, affecting overall user safety. The main difference between using the prediction algorithms was an increase in the amount of time the detection spent in the safe points, with EGP taking a longer time and hence better safety.

### B. COMBINED RESULTS: EFFICIENCY AND SAFETY

#### 1) EFFICIENCY

Results for efficiency in table 7 were presented as mean(sd). Overall, the prediction improved detection time and robot time with our implementation performing better than the baseline. This can be attributed to the augmentation of the velocity, which accounts for the improved model with less uncertainty, as shown in Figs 6 and 7. An improved time for detection resulted in improved robot time to reach the final destination which presented an overall improvement of 3%. The distance traveled by the robot remained the same across the models. This is because, despite more points detected, the robot didn't travel through all the intermediate points.

#### 2) USER SAFETY

Overall, results in table 8 showed that prediction with the STD(EGP) algorithm improved safety than the STC. This was due to a higher number of safe points traveled by the robot for the STD(EGP). The results can be attributed to the scaling factor in the STD(EGP) algorithm which improved the detection of the safe points. Another important factor to note is that the number of safe points travelled by the robot was lower than the detected safe points because the motion planning algorithm ignored most of the intermediate points detected. Safe points were located outside the user workspace. So by travelling through the safe points, the robot traveled a longer distance from the user. Results showed that with STD(EGP), the robot had a higher total distance from the user than the baseline, an improvement by 17%. Our implementation was safer for the human than the state-of-the-art.

**TABLE 7.** A table showing results of efficiency for strategy detection and robot motion.

| ST | $T_d$ | $T_r$ | $D_r$ |
| --- | --- | --- | --- |
| STA | 6.54(1.29) | 10.92(1.56) | 0.80 (0.02) |
| STB(EGP) | **6.36(1.28)** | **10.62(1.52)** | 0.80 (0.02) |

**TABLE 8.** A table showing results of safety for strategy detection and robot motion.

| ST | $SP_d$ | $SP_r$ | $D_h$ |
| --- | --- | --- | --- |
| STC | 1.86(0.37) | 0.57(0.19) | 0.82(0.06) |
| STD(EGP) | **2.57(0.17)** | **1.29(0.17)** | **0.96(0.03)** |

**TABLE 9.** Efficiency with eye gaze.

| ST | $T_d$ | $T_r$ | $D_r$ |
| --- | --- | --- | --- |
| STE | 6.66(1.35) | 10.15(1.28) | 1.30(0.21) |
| STF(EGP) | 6.80(1.34) | **9.99(1.04)** | 1.36(0.16) |

#### 3) EFFICIENCY WITH GAZE AND PREDICTION

With the addition of gaze for intention detection, Results in table 9 show that prediction along with gaze resulted in a shorter overall robot time.

Robot time was best for STF(EGP) at 9.99s, while without prediction, STE had the worst time at 10.15s. This shows that prediction improved robot time by 2%.

The distance traveled by the robot was the best with STF(EGP) at 1.36m compared to STE (1.30m), which was an indication that prediction required longer distances and time traveling through safe points that are safe and far from the human.

#### 4) SAFETY WITH GAZE AND PREDICTION

Results in table 10 showed that prediction improved safety. By analyzing the number of safe points detected, STF(EGP) had a higher number than STE. The number of safe points traveled through by the robot implied a larger the distance from the human. Overall, STF(EGP) gave the optimal value of distance which indicated an improvement of 13%. With the strategy, the robot traveled furthest from the user, implying that it was the safest.

### VII. DISCUSSION

In this study, we used a Gaussian process model to predict human trajectory motion in a virtual environment involving the use of a robot as a haptic interface with intermittent contacts. One of the significant challenges in human-robot interaction is developing systems that can interact with humans naturally and intuitively. This goes beyond integrating robots and humans with sensors but also developing algorithms for understanding and interpreting human intention.

**TABLE 10.** Safety with eye gaze.

| ST | $SP_d$ | $SP_r$ | $D_h$ |
|---|---|---|---|
| STE | 1.71(0.39) | 1.14(0.24) | 0.86(0.08) |
| STF(EGP) | **2.14(0.43)** | **1.43(0.19)** | **0.97(0.04)** |

Data-based machine learning models have been used to learn human motions and adapt the robot behavior to individual humans; however, transferability and adaptability to different use cases have been a problem mainly because many machine learning models require a lot of data to train, which may not be available in most cases. Secondly, training takes a lot of time and requires a lot of computational resources. This implies that for devices with low power and computation, such models are not feasible; the main solution is to train a model offline and use it online on a device for real-time inference. However, learning models need to perform better on unseen data. So for most applications, such models may not be effective for dynamic environments that often consist of unseen data. In this research, we address the above challenges first by demonstrating that it's possible to train a model online and then evaluate the training time to select suitable parameters necessary to achieve an optimized model for real-time training and inference.

We used efficient GP implementations and demonstrated that GP models can be used and trained online on small data to generate predictions for a real-time application. This research shows that GP implementation has a major bottleneck in matrix computation, and finding efficient methods for computing the kernel without inverting the matrix can greatly speed up the algorithm.

Much as many machine learning models have been used in prediction to adapt robot behavior during human-robot collaboration, most studies have evaluated prediction accuracy. Still, only some of them evaluate the efficiency regarding time to train and generate prediction because it has been offline. The research community has overlooked this yet it deserves attention when training is to be done online. This research addresses both the time and prediction accuracy of learning models.

Then we developed prediction strategies for human intention detection using GPs to predict a user trajectory online and used the strategies to improve device efficiency and safety in an immersive VR environment.

To improve the quality of predictions and efficiency for detection, we proposed a 2D GP to predict both the position and velocity. In general, our results showed that a our model had better accuracy compared to using the baseline in [27].

One major point to note is a lot of uncertainty characterizes that points not seen by the model in the training set. However, we were able to reduce it by using both positions and velocity to generate the final prediction. The results imply that multi-input GPs are better than single-input when the inputs are related. Lim et al. discovered such findings in their MOIHGP model [30]. Exploiting relationships in inter-related data is important for improved predictions with GP models.

To determine the user intention and move the robot to a user desired point, this study builds on the nearest neighbor algorithm proposed in [5] and [10] to introduce new strategies based on the GP prediction. Results showed improvement in detection and robot time against the state-of-the-art.

A critical aspect of any task where a human interacts with a robot is user safety. Previous studies showed that it was possible to improve safety by moving the robot as far away from the human as possible by using safe points; however, with a constant velocity, this would further result in robot delays, so they employed velocity modulation techniques to move the robot faster in between safe points to improve the speed of the robot [10]. This study builds on that by using GP prediction to implement new strategies that maximize the robot motion through safe points to move the robot to safety. However, in doing so, we had to sacrifice the time for the detection of interior point, and this slightly increased the robot's time. It's difficult to achieve both safety and efficiency simultaneously. With GP predictions in this study, we were able to achieve better results for both safety and robot time compared to the baseline. We further extend the strategies and take advantage of hand-eye coordination by using the human gaze to predict human intention and a GP to predict the hand motion. We were able to develop strategies that use a multi-modal interface to improve the human safety and efficiency of the robot. Results in table 10 show that with the GP predictions, both robot time and human safety were better than the state-of-the-art.

Another challenge is adaptability: Developing robots that can adapt to dynamic environments and tasks is still challenging. The strategies introduced in this article can be easily adapted to suit any task in human-robot interaction where prediction and teleoperation are required, including those outside virtual environments, because all the training and inference are done online and in real-time.

These strategies build on the work in [10] with a corresponding video illustration which explains and shows the strategies with the real hand motion in use for a real environment. Fig 14 shows the path of the real hand motion along with the predicted hand motion. To apply this method in the real environment, the same setup is used. Since both works use the same pre-recorded trajectories, the only adjustment is to use the points selected according to any of the strategies (STB, STD, or STF) which make use of the EGP prediction, and the robot simply executes the point-to-point trajectories resulting from the predicted hand motion.

To extend the technical evaluation done, in the future, we shall conduct user studies to assess other factors, such as human emotions and preferences for the different strategies. Secondly, our strategies used a combination of velocity positions to predict positions, and results showed that prediction by velocity was better than just positions alone. In the future, we would like to analyze the results when acceleration is used in the predictions. In addition, user safety was measured by distance from the user's body-centered in a virtual sphere. In the future, we would like to measure this from the whole body by adding more sensors to both the arms and reconstructing a full-body avatar. Another work would extend the experiments to many users and analyze their moods and perceptions using questionnaires.

## VIII. CONCLUSION

This paper aimed to improve the efficiency of a collaborative robot when used as a haptic interface with intermittent contacts while simultaneously improving the safety of the human user. We used Gaussian process models to predict human hand motion and developed strategies for human intention detection to improve safety and time for the robot to reach a desired point in a virtual environment. We then studied the effect of prediction using Gaussian process models based on a use case which is the analysis of the material of an automotive interior during the initial phases of interior design. Results from comparisons show that the Gaussian process prediction model improved efficiency and safety. The strategies have been presented separately to address different use cases for prediction, for applications that may use only hand tracking or a combination of both hand and gaze tracking. Unlike standard approaches that rely on large amounts of training data, Gaussian processes allow online training and prediction with small amounts of data, making them easy to use in more complex environments. This makes it suitable for real-time prediction tasks involving analysis of human body movements to determine human intent or when a human can influence robot motion, such as in robotic assistance, or teleoperation, and to improve safety and speed of equipment where human-robot contact is unavoidable.

## ACKNOWLEDGMENT
The authors would like to thank the reviewers for their helpful comments and suggestions.

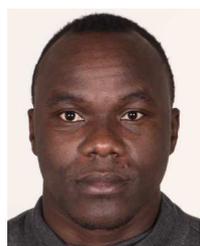
**STANLEY MUGISHA** received the B.S. and M.S. degrees in computer science from the University of Mysore, India, in 2017, and the Ph.D. degree in robotics and mechatronics from Universitàa di Genova, Italy, and École Centrale de Nantes, France, in 2022.

Since 2023, he has been a Post-Doctoral Research Fellow in human–robot collaboration with the University of Tartu, Estonia. His research interests include motion planning and control of manipulators, virtual reality, design, and development of safe systems for human–robot collaboration.

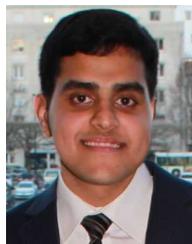
**VAMSI KRISHNA GUDA** received the B.Tech. degree in electronics and communications from SASTRA University, India, in 2016, and the M.S. degree in advanced robotics and the Ph.D. degree in collaborative robots from École Central de Nantes (ECN) in 2018 and 2022, respectively.

He is currently a Post-Doctoral Research Fellow in the research domain of robot control in human–robot interaction, focusing on robots for collaborative tasks at the Institut des Systèmes Intelligents et de Robotique (ISIR), Université Pierre-et-Marie-Curie, Paris. His research interests include trajectory planning and robot control of industrial robots (e.g., UR5 and Franka Emika), as well as developing prediction models for understanding human intentions, to enhance interactions between humans and robots.

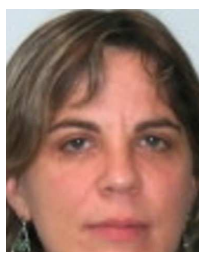
**CHRISTINE CHEVALLEREAU** received the master's and Ph.D. degrees from École Nationale Supérieure de Mécanique, Nantes, France, in 1985 and 1988, respectively. Since 1989, she has been with the CNRS, Institut de Recherche en Communications et Cybernétique de Nantes, and Laboratoire des Sciences du Numérique de Nantes (LS2N), École Central de Nantes, where she is currently the Deputy Director. Her research interests include modeling and control of manipulators and locomotor robots, in particular biped and bio-inspired robots.

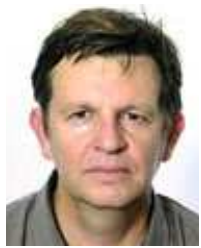
**DAMIEN CHABLAT** received the Ph.D. degree in mechanical engineering from École Central de Nantes and the University of Nantes, France, in 1998. Following a year at McGill University, he joined CNRS, in 1999, and became a Senior Researcher, in 2011. He is currently in robotics with Laboratoire des Sciences du Numérique de Nantes, École Central de Nantes, where he focuses on robotics, parallel manipulator design, and human fatigue evaluation. Since February 2023, he has led robotics initiatives for French National Research Agency (ANR). His skills and expertise include machine tools kinematics robotics design engineering automation and robotics product engineering mechatronics parallel robots and Maple (Software) among others.

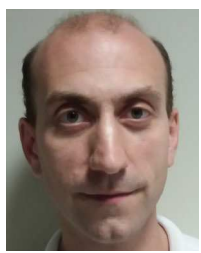
**MATTEO ZOPPI** (Member, IEEE) received the degree in mechanical engineering from the University of Genoa, in 2000, and the Ph.D. degree in robotics, in 2003. Since the end of 2014, he has been a Second Level University Professor with the Department of Mechanical Engineering, Università di Genova, where he is currently a Full Professor with the DIME-PMAR Robotics Group. His teachings include functional mechanical design, modeling of mechanical systems, mechanical design methods in robotics, and advanced modeling and simulation of mechanical systems. His scientific activities are also aimed at the methods of synthesis and analysis of mechanisms for robotics, screw theory in robotics, and design and development of robotic and automation systems.

• • •